\pdfoutput=1

\documentclass[11pt]{article}

\usepackage[]{acl}

\usepackage{times}
\usepackage{latexsym}

\usepackage[T1]{fontenc}

\usepackage[utf8]{inputenc}

\usepackage{microtype}
\usepackage{threeparttable}
\usepackage{graphicx}
\usepackage{float} 
\usepackage{subfigure}
\usepackage{caption}
\usepackage{bm}
\usepackage{multirow}
\usepackage{amsfonts}
\usepackage{amsmath}
\usepackage{enumitem}
\usepackage{arydshln}
\usepackage{algorithm}
\usepackage{algorithmic}
\title{Translatotron-V(ison): An End-to-End Model for \\ In-Image Machine Translation}

\author{
    Zhibin Lan\textsuperscript{1,3}\thanks{~~Work was done when Zhibin Lan was interning at Pattern Recognition Center, WeChat AI, Tencent Inc, China.}, 
    Liqiang Niu\textsuperscript{2}, 
    Fandong Meng\textsuperscript{2}, 
    Jie Zhou\textsuperscript{2}, 
    Min Zhang\textsuperscript{4}, 
    Jinsong Su\textsuperscript{1,3}\thanks{~~Corresponding author.}\\
    \textsuperscript{1}School of Informatics, Xiamen University, China\\
    \textsuperscript{2}Pattern Recognition Center, WeChat AI, Tencent Inc, China\\
    \textsuperscript{3}Key Laboratory of Digital Protection and Intelligent Processing of Intangible Cultural Heritage \\
    of Fujian and Taiwan (Xiamen University), Ministry of Culture and Tourism, China\\
    \textsuperscript{4}Institute of Computer Science and Technology, Soochow University, China\\
    \texttt{\small lanzhibin@stu.xmu.edu.cn
    }~~
    \texttt{\small jssu@xmu.edu.cn
    }
}

\begin{document}
\maketitle
\begin{abstract}
In-image machine translation (IIMT) aims to translate an image containing texts in source language into an image containing translations in target language. In this regard, 
conventional cascaded methods suffer from issues such as error propagation, massive parameters, and difficulties in deployment and 
retaining visual characteristics of the input image.
Thus, constructing end-to-end models has become an option, which, however, faces two main challenges: 1) the huge modeling burden, as it is required to simultaneously learn alignment across languages and preserve the visual characteristics of the input image; 2) the difficulties of directly predicting excessively lengthy pixel sequences.
In this paper, we propose \textit{Translatotron-V(ision)}, an end-to-end IIMT model consisting of four modules. In addition to an image encoder, and an image decoder, our model contains a target text decoder and an image tokenizer. Among them, the target text decoder is used to alleviate the language alignment burden, and the image tokenizer converts long sequences of pixels into shorter sequences of visual tokens, preventing the model from focusing on low-level visual features. Besides, we present a two-stage training framework for our model to assist the model in learning alignment across modalities and languages. Finally, we propose a location-aware evaluation metric called Structure-BLEU to assess the translation quality of the generated images. Experimental results demonstrate that our model achieves competitive performance compared to cascaded models with only 70.9\% of parameters, and significantly outperforms the pixel-level end-to-end IIMT model.\footnote{Our code and dataset can be found at \url{https://github.com/DeepLearnXMU/translatotron-v}.}
\end{abstract}

\section{Introduction}
In recent years, significant advancements have been achieved in natural language processing (NLP) and computer vision (CV), largely due to the evolution of deep learning. As a combined direction of these two fields, in-image machine translation (IIMT) aims to covert an image containing texts in source language into another image containing the translations in target language, which has significant research value and practical applications. It not only helps us understand the fusion mechanism of multimodal and multilingual information, but also finds widespread applications in daily life. For instance, IIMT can effortlessly enable foreign travelers to read signs written in other languages. 

\begin{figure}[]
\setlength{\abovecaptionskip}{-1pt}
\setlength{\belowcaptionskip}{-0.3cm}
    \centering
    \includegraphics[width=1\linewidth]{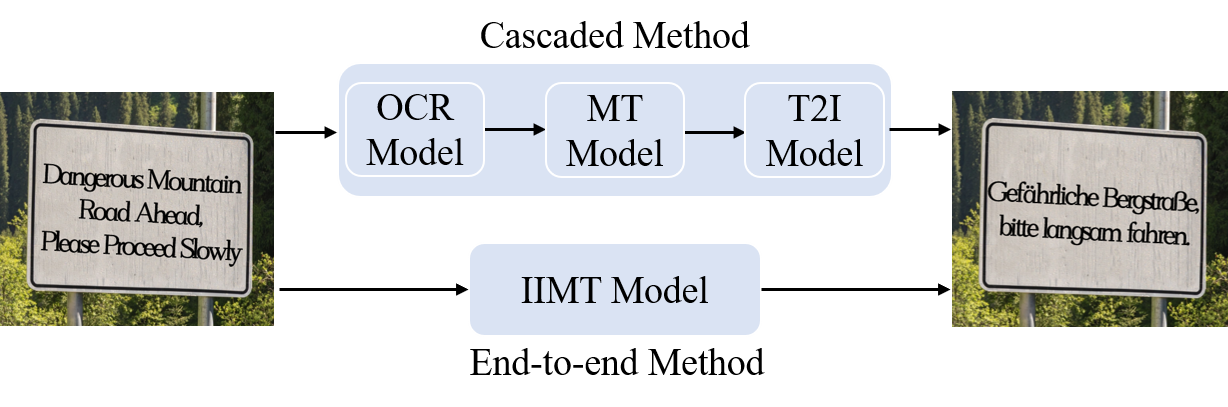}
\caption{The illustration of two paradigms of IIMT.}
\label{figure:task_example}
\end{figure}

As shown in Figure \ref{figure:task_example}, current IIMT systems are divided into two paradigms: cascaded and end-to-end. The first one relies on cascading multiple models, including an optical character recognition (OCR) model, a machine translation (MT) model, and a text-to-image (T2I) model. However, this paradigm suffers from error propagation, massive parameters, and difficulties in deployment and retaining visual characteristics of the input image.
By contrast, end-to-end methods \cite{DBLP:journals/corr/abs-2010-10648,DBLP:conf/emnlp/TianLLGW23} integrate different models into one IIMT model and conduct end-to-end training.
Thus, they have potential advantages over cascaded systems in aspects of avoiding error propagation, reduced parameters, and ease of deployment.
Particularly, they are naturally capable of retaining visual characteristics from the input image during translation, e.g. maintaining background, text location, font, etc.

Despite the above advantages, the end-to-end IIMT models still face two major challenges: 1) the huge modeling burden, since they are required to not only learn alignment between two languages but also the visual characteristics of the input image; 2) the difficulties of directly predicting excessively lengthy pixel sequences, which are low-level and involve a large search space \cite{DBLP:conf/icml/RameshPGGVRCS21, DBLP:journals/tmlr/YuXKLBWVKYAHHPLZBW22}.

To the best of our knowledge, \cite{DBLP:journals/corr/abs-2010-10648} and \cite{DBLP:conf/emnlp/TianLLGW23} are the only two attempts to explore end-to-end IIMT. However, the former is directly based on pixel prediction, resulting in significantly lower translation quality compared to the cascaded models, while the latter requires converting RGB images to grayscale ones, losing visual characteristics. Besides, both of them can only handle images containing single-line text. These defects make them still far from real-world applications.

\begin{figure*}[]
\setlength{\abovecaptionskip}{-0.5pt}
    \centering
    \includegraphics[width=0.90\linewidth]{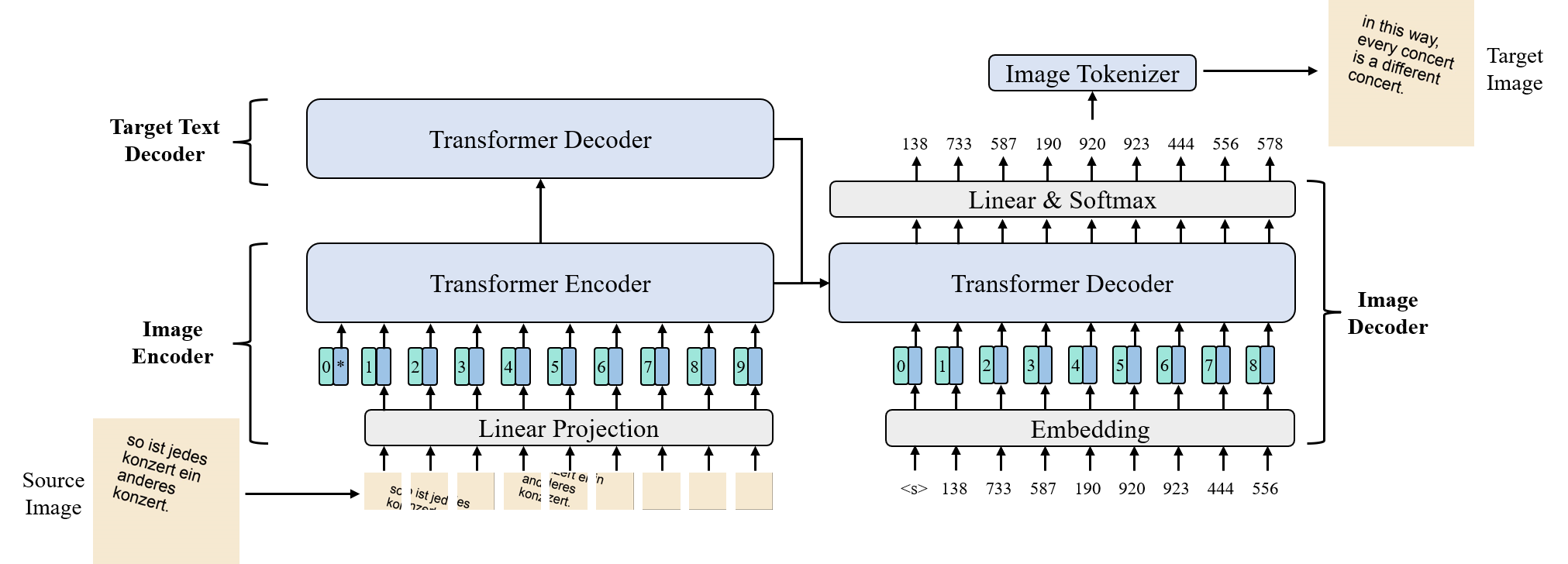}
\caption{Overview of our model.}

\label{figure:model}
\vspace{-0.3cm}
\end{figure*}

In this paper, we propose \textit{Translatotron-V(ision)}, the first end-to-end IIMT model capable of generating RGB images, achieving comparable performance to cascaded models with only 70.9\% of parameters. As shown in Figure \ref{figure:model}, our model consists of four modules: 1) an \textit{image encoder} that represents the semantics of the image as a sequence of visual vectors; 2) a \textit{target text decoder} that utilizes the visual vector sequence to predict the text translation, which can effectively reduce the modeling burden on the image decoder; 3) an \textit{image decoder} that generates the visual tokens of the target image based on the visual and linguistic information generated from the image encoder and target text decoder, respectively; 4) an \textit{image tokenizer} that converts the image into discrete visual tokens and can reconstruct the image from these visual tokens. By converting the image into visual tokens, the image decoder only needs to predict visual tokens, rather than excessively lengthy pixel sequences, which allows the model to avoid spending too much capacity capturing low-level visual features.

Furthermore, as illustrated in Figure \ref{figure:training}, we propose a training framework for our model, consisting of two stages. First, we utilize large-scale unlabeled images to train the image tokenizer through an image reconstruction task. Then, we freeze the image tokenizer and train other modules using IIMT dataset. Inspired by end-to-end speech translation \cite{DBLP:conf/interspeech/JiaWBMJCW19}, we introduce multi-task learning at this stage. The auxiliary tasks include OCR and text image translation (TIT), assisting the model in learning alignment across different modalities and languages. Particularly, we introduce a knowledge distillation method to reduce the difficulty of the end-to-end model directly learning from ground-truth labels.

Due to the absence of publicly available IIMT datasets, we use IWSLT14 German-English \cite{DBLP:conf/iwslt/CettoloNSBF14} to synthesize a dataset for this task. Note that unlike previous works \cite{DBLP:journals/corr/abs-2010-10648, DBLP:conf/emnlp/TianLLGW23} only focus on images containing single-line text, the images in our dataset are more complex, featuring multiple lines of text, as well as text rotation and translation. Furthermore, since the conventional BLEU \cite{DBLP:conf/acl/PapineniRWZ02} is not applicable to image evaluation, we extend BLEU to Structure-BLEU that considers text location information to better evaluate the quality of text translations within images.

To summarize, we have the following major contributions in this work:
\begin{itemize}[topsep=0pt,itemsep=0pt,parsep=0pt]
\item We propose a novel end-to-end IIMT model named Translatotron-V. More importantly, it introduces two crucial modules to address major challenges in end-to-end IIMT: 1) target text decoder used to alleviate the modeling burden; 2) image tokenizer preventing the model from directly predicting pixels.

\item We present a two-stage training framework for Translatotron-V, which fully exploits unlabeled images, OCR, and TIT data to refine the model training.

\item We propose Structure-BLEU, an evaluation metric that considers text location information for IIMT.

\item Experimental results demonstrate that Translatotron-V not only significantly outperforms the pixel-level end-to-end IIMT model, but also achieves comparable performance with fewer parameters to cascaded models.
\end{itemize}

\section{Related Work}
To achieve high-performance IIMT, previous research mainly focuses on text image translation (TIT), which is a subtask of IIMT \cite{DBLP:conf/icpr/WatanabeOKT98, DBLP:conf/icassp/YangCZZW02, DBLP:conf/icdar/DuHSS11, DBLP:journals/ijdar/ChenCN15, DBLP:conf/lt4dh/AfliW16,DBLP:conf/acl/LanYLZ0WHS23}. Unlike conventional multimodal machine translation \cite{DBLP:conf/acl/ElliottFSS16,DBLP:conf/acl/YinMSZYZL20,DBLP:conf/mm/LinMSYYGZL20,DBLP:journals/isci/SuCJZLGWL21,DBLP:journals/ai/YinZSZMZHL23,DBLP:liyan}, TIT aims to translate source language texts in images into target language. In this regard, dominant studies resort to the cascading method, which uses an OCR model to obtain the recognized source language texts and then feed them into an MT model for translation \cite{DBLP:journals/corr/GoodfellowBIAS13,DBLP:conf/emnlp/ZhangXSDZ16,DBLP:conf/iclr/Gu0XLS18}. 

Afterwards, due to the advantages of mitigating error propagation, the end-to-end TIT attracts increasing attention. 
\citeauthor{DBLP:conf/icpr/ChenYZYL20} (\citeyear{DBLP:conf/icpr/ChenYZYL20}) adopt multi-task learning framework that integrates OCR as an auxiliary task. Along this line, \citeauthor{DBLP:conf/icpr/MaZTHWZ022} (\citeyear{DBLP:conf/icpr/MaZTHWZ022}) incorporating MT into the multi-task learning framework. Unlike previous studies, both \citeauthor{DBLP:conf/icdar/SuLZ21} (\citeyear{DBLP:conf/icdar/SuLZ21}) and \citeauthor{DBLP:journals/corr/abs-2305-05166} (\citeyear{DBLP:journals/corr/abs-2305-05166}) employ an adapter to combine individual pretrained OCR and MT modules in a TIT model. Furthermore, \citeauthor{DBLP:journals/corr/abs-2305-05226} (\citeyear{DBLP:journals/corr/abs-2305-05226}) apply knowledge distillation to effectively distillate the knowledge of OCR and MT models into the end-to-end TIT model. \citeauthor{DBLP:conf/acl/ZhuLLX23} (\citeyear{DBLP:conf/acl/ZhuLLX23}) explore an end-to-end TIT model with an aligner and a regularizer to reduce the modality gap. To explicitly exploit guidance from recognized texts, \citeauthor{DBLP:conf/emnlp/MaZTZZZ23} (\citeyear{DBLP:conf/emnlp/MaZTZZZ23}) incorporate recognized text information into the TIT decoder through interactive attention. Differing from the above studies focusing on model design, \citeauthor{DBLP:conf/emnlp/SaleskyEP21} (\citeyear{DBLP:conf/emnlp/SaleskyEP21}) analyze the effect of visual text representation, and find that it exhibits significant robustness to various types of noise.

However, none of the aforementioned works consider generating the image with target translations, which is a common requirement in real-world scenarios. To this end, \citeauthor{DBLP:journals/corr/abs-2010-10648} (\citeyear{DBLP:journals/corr/abs-2010-10648}) first explore the IIMT task.  They introduce an end-to-end model that contains a self-attention encoder, two convolutional encoders, and a convolutional decoder to generate target images at the pixel level. Nonetheless, their model significantly lags behind cascaded models, suffering from issues such as character omission and artifacts. Recently, \citeauthor{DBLP:conf/emnlp/TianLLGW23} (\citeyear{DBLP:conf/emnlp/TianLLGW23}) convert pixels into characters, thereby transforming the IIMT task into a conventional sequence-to-sequence text generation task. However, this method can only generate grayscale images, losing visual characteristics. \citeauthor{DBLP:journals/corr/abs-2308-03024} (\citeyear{DBLP:journals/corr/abs-2308-03024}) present a conditional diffusion-based image editing model, which replaces text in the input image with a given translation while preserving the visual characteristics of origin image. However, this model can only perform single-word editing, which makes its application very limited. 

Different from these studies, we propose an end-to-end IIMT model that can generate RGB images with multiple lines of text while preserving the visual features of the input image, and achieve comparable performance to the cascaded model.

\section{Our Model}

\subsection{Model Architecture}
\label{section:3.1}
As shown in Figure \ref{figure:model}, our model consists of four modules: an image encoder, a target text decoder, an image decoder, and an image tokenizer. All of those modules will be elaborated in the following.

\textbf{Image Encoder}. This module converts the input image into a sequence of visual vectors.

We use ViT \cite{DBLP:conf/iclr/DosovitskiyB0WZ21} as the backbone of the image encoder. In order to convert a 2D image into a 1D sequence that can be handled by Transformer, we first split the input image $\mathbf{x}$ into $N=HW/P^2$ image patches $\{x_{i}\}_{i=1}^N$, where $(H,W)$ is the resolution of the input image, and $(P,P)$ is the resolution of each patch. Then we apply a linear projection matrix $\mathbf{W}_e$ to transform image patches into patch embeddings, and use a standard learnable positional embedding matrix $\mathbf{E}_{pos}$ to further optimize these patch embeddings. Formally, the initial hidden states $\mathbf{H}_{ie}^0$ of the image encoder can be formulated as
\setlength{\abovedisplayskip}{3pt}
\begin{equation}
    \mathbf{H}_{ie}^{(0)} = [x_0;\mathbf{W}_ex_1;\mathbf{W}_ex_2;...;\mathbf{W}_ex_N] + \mathbf{E}_{pos},
\end{equation}
where $x_0$ is the special token prepended to the input sequence.

Afterwards, we process these patch embeddings using a Transformer encoder with multiple layers. Each Transformer encoder layer is composed of a self-attention sub-layer and a feed-forward network (FFN) sub-layer. Layernorm (LN) is applied before each sub-layer, and residual connections after each sub-layer \cite{DBLP:conf/acl/WangLXZLWC19}. The hidden states $\mathbf{H}_{ie}^{(l)}$ of the $l$-th encoder layer is calculated as
\setlength{\abovedisplayskip}{3pt}
\begin{equation}
\mathbf{H}_{ie}^{(l)}=\mathrm{FFN}(\mathrm{MHA}(\mathbf{H}_{ie}^{(l-1)},\mathbf{H}_{ie}^{(l-1)},\mathbf{H}_{ie}^{(l-1)})),
\end{equation}
where $\mathrm{MHA}(\cdot,\cdot,\cdot)$ denotes a multi-head attention function. The residual connection and layer normalization are omitted for simplicity.

\textbf{Target Text Decoder}. By utilizing the features generated by the image encoder, this decoder is responsible for producing text translations. In this way, it focuses on the alignment of different languages, and thus alleviates the modeling burden of the image decoder.

When constructing our target text decoder, we employ the widely-used Transformer \cite{DBLP:conf/nips/VaswaniSPUJGKP17} decoder as the architecture, consisting of multiple identical layers. In addition to the standard self-attention and FFN sub-layers, each decoder layer is equipped with a cross-attention sub-layer to exploit hidden states produced by the image encoder. Formally, we calculate the hidden states $\mathbf{H}_{td}^{(l)}$ for the $l$-th decoder layer using the following equations:
\setlength{\abovedisplayskip}{3pt}
\setlength{\belowdisplayskip}{3pt}
\begin{align}
\mathbf{C}_{td}^{(l)}&=\mathrm{MHA}(\mathbf{H}_{td}^{(l-1)},\mathbf{H}_{td}^{(l-1)},\mathbf{H}_{td}^{(l-1)}), \\
\mathbf{H}_{td}^{(l)}&=\mathrm{FFN}({\mathrm{MHA}(\mathbf{C}_{td}^{(l)},
\mathbf{H}_{ie}^{(L)},\mathbf{H}_{ie}^{(L)}))},
\end{align}
where the initial hidden states $\mathbf{H}_{td}^{(0)}$ are computed by summing the word embeddings and position embeddings of the input sequence. Unless otherwise specified, we use $L$ to represent the last layer.

\textbf{Image Decoder}. This module is responsible for generating visual tokens based on visual and linguistic information generated from the image encoder and target text decoder, respectively. 

The architecture of the image decoder closely resembles that of the target text decoder but with the following notable modifications. It includes two cross-attention sub-layers to gather information from both the image encoder and target text decoder, followed by a fusion sub-layer to generate intermediate representations enriched with both visual and linguistic features. Besides, we incorporate the 2D relative position encoding \cite{DBLP:conf/iccv/WuPCFC21} into the self-attention sub-layer to capture relative positional relationships within images.

Let $\mathbf{C}_{id}^{(l)}$ denote the hidden states output by the $l$-th self-attention sub-layer, we calculate it in the following way:
\begin{equation}
\mathbf{C}_{id}^{(l)}=\mathrm{MHA}(\mathbf{H}_{id}^{(l-1)},\mathbf{H}_{id}^{(l-1)},\mathbf{H}_{id}^{(l-1)}),
\end{equation}
where $\mathbf{H}_{id}^{(l-1)}$ represents the hidden state output by the ($l$-1)-th image decoder layer. Subsequently, the hidden states $\mathbf{\overline{H}}_{id}^{(l)}$ and $\mathbf{\widetilde{H}}_{id}^{(l)}$ are computed through two cross-attention mechanisms, which attend to the image encoder and the target text decoder, respectively, as follows:
\setlength{\abovedisplayskip}{3pt}
\setlength{\belowdisplayskip}{2pt}
\begin{equation}
\mathbf{\overline{H}}_{id}^{(l)}=\mathrm{MHA}(\mathbf{C}_{id}^{(l)},\mathbf{H}_{ie}^{(L)},\mathbf{H}_{ie}^{(L)}),
\end{equation}
\begin{equation}
\mathbf{\widetilde{H}}_{id}^{(l)}=\mathrm{MHA}(\mathbf{C}_{id}^{(l)},\mathbf{H}_{td}^{(L)},\mathbf{H}_{td}^{(L)}).
\end{equation}

Finally, the hidden states of the $l$-th image decoder layer are obtained through a gated fusion mechanism, which is calculated using the following equations:
\begin{align}
    \Lambda &= \mathrm{sigmoid}(\mathbf{W}_{\Lambda}\mathbf{\overline{H}}_{id}^{(l)}+\mathbf{U}_{\Lambda}\mathbf{\widetilde{H}}_{id}^{(l)}), \\
    \mathbf{H}_{id}^{(l)} &= \Lambda \mathbf{\overline{H}}_{id}^{(l)} + (1-\Lambda)\mathbf{\widetilde{H}}_{id}^{(l)},
\end{align}
where $\mathbf{W}_{\Lambda}$ and $\mathbf{U}_{\Lambda}$ are projection matrices, and $\Lambda$ is a gated matrix featuring values ranging from 0 to 1, serving the purpose of dynamically fusing two modalities of information.

\textbf{Image Tokenizer}. It is used to perform the conversion between an image and a sequence of discrete visual tokens. By introducing this module, we allow the image decoder only to predict visual tokens, preventing it from modeling excessively lengthy sequences. For instance, a 256$\times$256$\times$3 RGB image results in 196,608 rasterized values. 

Our image tokenizer follows the architecture of ViT-VQGAN \cite{DBLP:conf/iclr/YuLKZPQKXBW22}, which includes a Vison Transformer (ViT) \cite{DBLP:conf/iclr/DosovitskiyB0WZ21} based encoder and decoder. The encoder $E$ of the image tokenizer is used to tokenize the image into $\mathbf{z}=(z_1,...,z_N)$ through a quantizer $q(\cdot)$. Formally, the quantizer looks up the nearest visual token for each input, as shown in the following:
\begin{equation}
    z_i = q(E(x_i)) =\mathop{\mathrm{argmin}}\limits_{e_{k}\in \mathcal{V}}||E(x_i)-e_{k}||_2,
    \label{quantization}
\end{equation}
where $\mathcal{V}$ is the image vocabulary containing visual tokens.

Conversely, the decoder $G$ of the image tokenizer reconstructs the input image based on the visual tokens generated by $E$, formulated as
\setlength{\abovedisplayskip}{3pt}
\setlength{\belowdisplayskip}{3pt}
\begin{equation}
    \hat{\mathbf{x}} = G(q(E(\mathbf{x}))).
\end{equation}

Please note that during training, we use the encoder to obtain visual tokens of the target image as labels. During inference, the decoder converts visual tokens generated by the image decoder into the target image.

\begin{figure*}[]
    \centering
    \includegraphics[width=0.85\linewidth]{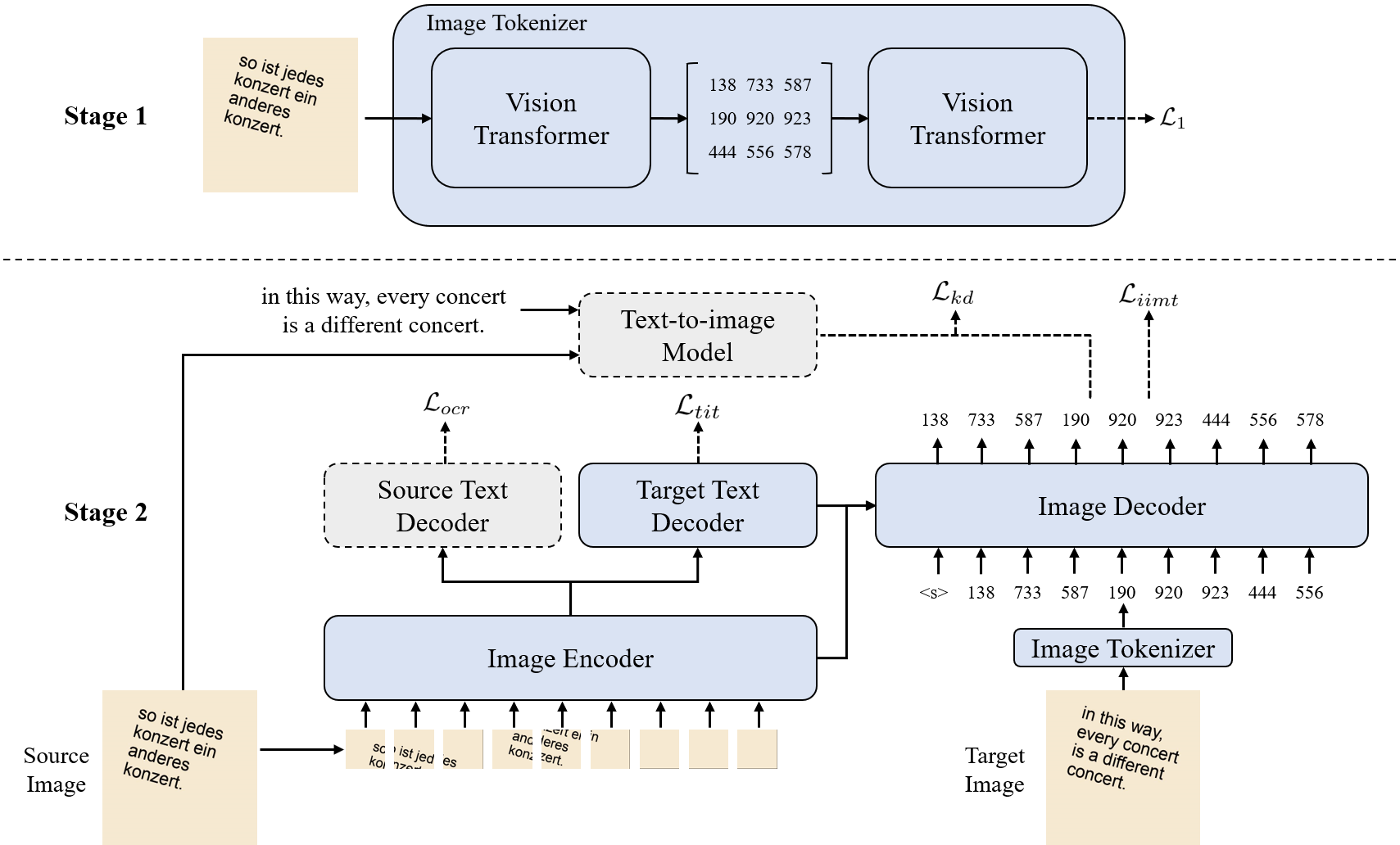}
\caption{Overview of our two-stage training framework. The
modules enclosed by dotted lines will be removed during inference.}

\label{figure:training}
\vspace{-0.3cm}
\end{figure*}

\subsection{Model Training}
We provide a detailed description of the training procedures for our model, which consists of two stages, as illustrated in Figure \ref{figure:training}.

\textbf{Stage 1}. At this stage, we train the image tokenizer using a large-scale unlabeled image dataset $D_v$ in the same way as ViT-VQGAN \cite{DBLP:conf/iclr/YuLKZPQKXBW22}, where we convert the input image into visual tokens and then reconstruct the image from these visual tokens.

Given an image $\mathbf{x}$ from the unlabeled image dataset $D_{v}$, we define the training objective of this stage as follows:
\begin{equation}
\begin{aligned}
    \mathcal{L}_{1} = ||\hat{\mathbf{x}}-\mathbf{x}||^2 &+ 
  ||\mathrm{sg}(E(\mathbf{x}))-\mathbf{z}||_2^2 \\ &+ \beta||E(\mathbf{x})-\mathrm{sg}(\mathbf{z})||_2^2.
\end{aligned}
\end{equation}
Here, the first item is the reconstruction loss optimizing the encoder and decoder, the middle item is the vector-quantization loss used to update the visual tokens, the last item is the so-called ``commitment loss''  for the encoder which prevents its output fluctuating frequently from one visual token to another, $\mathrm{sg}(\cdot)$ denotes the stop-gradient operation, and $\beta$ is the weighting factor set to 0.25 following \citeauthor{DBLP:conf/nips/OordVK17} (\citeyear{DBLP:conf/nips/OordVK17}).\footnote{Note that we also include other loss terms as presented in \cite{DBLP:conf/iclr/YuLKZPQKXBW22}, but omit the descriptions for brevity. Please refer to \cite{DBLP:conf/iclr/YuLKZPQKXBW22} for more details.}


\textbf{Stage 2}. Using an IIMT dataset, we then adopt multi-task learning and knowledge distillation to train the image encoder, target text decoder, and image decoder.

Overall, the training objective at this stage is defined as follows:
\begin{equation}
    \mathcal{L}_2 = \mathcal{L}_{iimt} + \mathcal{L}_{ocr} + \mathcal{L}_{tit} + \mathcal{L}_{kd}.
\end{equation}
where $\mathcal{L}_{iimt}$, $\mathcal{L}_{ocr}$, $\mathcal{L}_{tit}$, and $\mathcal{L}_{kd}$ denote the IIMT task loss, OCR auxiliary task loss, TIT auxiliary task loss, and knowledge distillation loss, respectively. \footnote{We also explore the balance of different training objectives. Experimental results in Appendix \ref{appendix:weight_balance} show that we do not need to introduce additional hyperparameters to balance different objectives.}

Given an IIMT training instance $(\mathbf{x},\mathbf{y},\mathbf{s},\mathbf{t})$ from the IIMT dataset $D_{iimt}$, we can utilize the image tokenizer trained in the first stage to process the target image, obtaining visual tokens denoted as $\mathbf{z}$. Here, $\mathbf{x}$ represents the source image, $\mathbf{y}$ is the target image, $\mathbf{s}$ denotes the source language text within the source image, and $\mathbf{t}$ denotes the target language text within the target image.

To alleviate the burden of end-to-end model training, we adopt multi-task learning, which involves not only the primary IIMT task but also two auxiliary tasks: the OCR task and the TIT task. The OCR task is employed to assist the model in recognizing texts within the image, while the TIT task further facilitates cross-lingual alignment. Formally, the training objective of the IIMT task can be formulated as follows:
\begin{equation}
    \mathcal{L}_{iimt}=-\mathrm{log}p(\mathbf{z}|\mathbf{x};\theta_{ie},\theta_{ttd},\theta_{id}),
\end{equation}
where $\theta_{ie}$, $\theta_{ttd}$, $\theta_{id}$ denote the trainable parameters of the \textbf{i}mage \textbf{e}ncoder, \textbf{t}arget \textbf{t}ext \textbf{d}ecoder, and \textbf{i}mage \textbf{d}ecoder, respectively.

To train our model using the OCR auxiliary task, we additionally introduce a source text decoder, which adopts the same architecture as the target text decoder. Formally, the training objectives of the OCR and TIT auxiliary tasks are defined as
\begin{equation}
    \mathcal{L}_{ocr}=-\mathrm{log}p(\mathbf{s}|\mathbf{x};\theta_{ie},\theta_{std}),
\end{equation}
\begin{equation}
    \mathcal{L}_{tit}=-\mathrm{log}p(\mathbf{t}|\mathbf{x};\theta_{ie},\theta_{ttd}),
\vspace{-3pt}
\end{equation}
where $\theta_{std}$ is the parameters of the \textbf{s}ource \textbf{t}ext \textbf{d}ecoder. Note that the source text decoder takes the intermediate hidden states of the image encoder as input. This design is based on the intuition that the shallow encoder layers represent the source visual content, while the deep layers encode more information about the target visual content.

Besides, training an end-to-end model is considerably more difficult than a T2I model, where the latter only needs to learn the mapping between different modalities and thus has better performance. Consequently, we introduce a T2I model as a teacher to facilitate knowledge transfer to the end-to-end model. This T2I model includes a Transformer-based text encoder, a ResNet-based image encoder \cite{DBLP:conf/cvpr/HeZRS16}, and an image decoder similar to our model, where the image encoder is used to preserve the features of the original image.
Denote the output distribution of the teacher model for $t$-th visual token $z_t$ as $q(z_t|\mathbf{z}_{<t},\mathbf{x},\mathbf{t};\theta_{t2i})$, we define the cross-entropy between the distributions of teacher and student as the distillation loss:
\begin{equation}
    \begin{split}
    \mathcal{L}_{kd} = &-\sum_{t=1}^{|\mathbf{z}|}\sum_{k=1}^{|\mathcal{V}|}q(z_t=k|\mathbf{z}_{<t},\mathbf{x},\mathbf{t};\theta_{t2i}) \\
    &\mathrm{log}p(z_t=k|\mathbf{z}_{<t},\mathbf{x};\theta_{ie}, \theta_{ttd}, \theta_{id}),
    \end{split}
\end{equation}
where $\theta_{t2i}$ represents the parameters of the T2I model.
Note that we will remove the source text decoder and T2I model during inference.

\section{Experiments}
\subsection{Setup}

\textbf{Dataset}. Due to the lack of readily available data, we utilize the widely-used IWSLT14 German-English (De-En) dataset \cite{DBLP:conf/iwslt/CettoloNSBF14} to synthesize paired images for this task. Concretely, we leverage the Python Pillow package\footnote{https://github.com/python-pillow/Pillow} to render texts onto images with the black Arial\footnote{https://learn.microsoft.com/en-us/typography/font-list/arial} font. The text is arranged horizontally from left to right, and vertically from top to bottom, with randomly translating and rotating. This involves shifting the text in a random direction and changing its orientation by a random angle. Additionally, the background color of the image is selected randomly and the resolution of the images is 512$\times$512. Note that bilingual texts exceeding the image boundaries will be disregarded during the process of data synthesis. In contrast to prior studies \cite{DBLP:journals/corr/abs-2010-10648, DBLP:conf/emnlp/TianLLGW23}, which focus solely on generating images with single-line text and white background, our research delves into more complex scenes. In the end, the synthesized dataset comprises 81,741 training instances, 3,765 validation instances, and 3,527 test instances. Several synthetic examples and comparisons with previous data can be found in Appendix \ref{appendix:data_examples}.

In addition to the IIMT data, a substantial quantity of images is also indispensable for training the image tokenizer. To this end, we employ the text extracted from the WMT14 English-German (En-De) \cite{ws-2014-statistical} to synthesize images for training our image tokenizer.

\begin{table*}[]
\centering

\resizebox{2\columnwidth}{!}{
\begin{tabular}{lccccccc}
\hline
\multicolumn{1}{l|}{\multirow{2}{*}{\textbf{Model}}}                    & \multicolumn{3}{c|}{\textbf{De$\rightarrow$En}}                                                                                                     & \multicolumn{3}{c|}{\textbf{En$\rightarrow$De}}                                                                                                    & \multirow{2}{*}{\#Param}       \\
\multicolumn{1}{l|}{}                                          & BLEU $\uparrow$    & Structure-BLEU $\uparrow$ & \multicolumn{1}{c|}{SSIM $\uparrow$}   & BLEU $\uparrow$    & Structure-BLEU $\uparrow$ & \multicolumn{1}{c|}{SSIM $\uparrow$}   &                                \\ \hline
\multicolumn{8}{c}{\textit{Cascaded Models}}                                                                                                                                                                                                                                                                                                                                 \\ \hline
\multicolumn{1}{l|}{OCR+MT+T2I}                            & 15.37                           & 14.87                                  & \multicolumn{1}{c|}{0.7785}                           & 13.22                           & 12.57                                  & \multicolumn{1}{c|}{0.7550}                           & 247M                           \\ 
\multicolumn{1}{l|}{TIT+T2I}                                 & 14.80                           & 14.73                                  & \multicolumn{1}{c|}{0.7812}                           & 12.92                           & 12.74                                  & \multicolumn{1}{c|}{0.7620} & 201M                           \\
\multicolumn{1}{l|}{PEIT+T2I}                                 & 10.91                           & 10.78                                  & \multicolumn{1}{c|}{0.7740}                           & 8.77                           & 8.01                                  & \multicolumn{1}{c|}{0.7594} & 178M                           \\
\hline
\multicolumn{8}{c}{\textit{End-to-end Models}}                                                                                                                                                                                                                                                                                                                                \\ \hline
\multicolumn{1}{l|}{Pixel-level Transformer} & 0.15 & 0.15        & \multicolumn{1}{c|}{0.7538} &1.11  &1.22         & \multicolumn{1}{c|}{0.7616}                           & 162M \\
\hdashline[2pt/2pt]
\multicolumn{1}{l|}{Translatotron-V} & \textbf{15.39} & \textbf{15.26}        & \multicolumn{1}{c|}{\textbf{0.7832}} & \textbf{13.23} & \textbf{12.92}        & \multicolumn{1}{c|}{\textbf{0.7629}}                           & 175M \\ \hline
\end{tabular}
}
\caption{Experimental results on the De$\rightarrow$En and En$\rightarrow$De IIMT tasks.}
\label{tab:de_en_results}
\end{table*}

\begin{table}[]
\centering
\resizebox{1\columnwidth}{!}{
\begin{tabular}{l|ccc}
\hline
\textbf{Model}      & \textbf{BLEU} $\uparrow$          & \textbf{S-BLEU} $\uparrow$  & \textbf{SSIM} $\uparrow$\\
\hline
Translatotron-V      & \textbf{15.39} & \textbf{15.26}          & \textbf{0.7832}  \\
\hline
\hspace{1em}\textit{w/o} gated fusion   &14.34  &14.20       & 0.7830       \\
\hspace{1em}\textit{w/o} OCR auxiliary task &1.39  & 1.18       &0.7277     \\
\hspace{1em}\textit{w/o} knowledge distillation   &13.35        & 13.43      &0.7813      \\
\hspace{1em}\textit{w/o} target text decoder     &0.47  & 0.43      &0.7751       \\
\hline
\end{tabular}
}
\caption{Ablation study of Translatotron-V on the De$\rightarrow$En IIMT task. S-BLEU represents Structure-BLEU.}
\label{tab:ablation_study}
\end{table}

\textbf{Implementation Details}.
In this work, we employ the same setting as ViT-B \cite{DBLP:conf/iclr/DosovitskiyB0WZ21} to construct our image encoder. Both our target text decoder and image decoder are composed of 8 layers, each of which has 512-dimensional hidden states, 8 attention heads, and 2,048 feed-forward hidden states. Besides, our image tokenizer is similar to ViT-VQGAN-SS \cite{DBLP:conf/iclr/YuLKZPQKXBW22} but uses a smaller setup. It consists of 4 layers of encoder and decoder, each of which has 256-dimensional hidden states, 8 attention heads, and 1,024 feed-forward hidden states. Particularly, we use a character-level vocabulary of size 256 for the OCR and TIT auxiliary tasks, while the image vocabulary size for visual tokens is set to 8,192. Unless otherwise specified, the patch size of the image is set to 16.

During the first training stage, we train the image tokenizer with a batch size of 512 for 10,000 steps, where the parameters are updated by AdamW \cite{DBLP:conf/iclr/LoshchilovH19} with $\beta_1=0.9, \beta_2=0.99$. During the second stage, we train the model for 100 epochs with an early stopping patience set to 10, and a batch size set to 80. This stage of training also utilizes the AdamW optimizer ($\beta_1=0.9, \beta_2=0.999$) along with weight decay of 0.001 and polynomial decay learning rate scheduling. To alleviate overfitting, we apply a dropout rate of 0.1 and incorporate the label smoothing with a coefficient of 0.1, and we average the checkpoints of the last 10 epochs for evaluation.

\textbf{Baselines}.
We construct the following baselines: OCR+MT+T2I, TIT+T2I, PEIT+T2I, and pixel-level Transformer, all models trained using character inputs and outputs similar to our model.

1) \textit{OCR+MT+T2I}. This baseline cascades three models: an OCR model, an MT model, and a T2I model. Note that our teacher model has the same architecture as the T2I Model, except for reducing the hidden states from 512-dimensional to 384-dimensional. 2) \textit{TIT+T2I}. We construct this baseline by cascading the TIT model and the T2I model.
Additionally, it applies both OCR and TIT tasks during training to achieve better performance. 3) \textit{PEIT+T2I}. This baseline is similar to TIT+T2I but replaces the multi-line TIT model with PEIT \cite{DBLP:conf/acl/ZhuLLX23}, the state-of-the-art single-line TIT model. Since PEIT is designed for single-line TIT, during inference, we first employ the widely-used EasyOCR\footnote{https://github.com/JaidedAI/EasyOCR} as the detection model to recognize and crop each line of text from the image, and then concatenate them together into a single-line text image. 4) \textit{Pixel-level Transformer}. This model uses the same structure as our model but removes the image tokenizer and directly predicts pixel values. It is trained using multi-task learning as well, with the IIMT task being optimized with a mean squared error loss due to the pixel values being of floating-point type. 

The detailed architecture of OCR+MT+T2I, TIT+T2I, and PEIT+T2I is described in Appendix \ref{appendix:baseline_details}.

\textbf{Evaluation}.
We evaluate the output images from both the perspectives of translation quality and image quality. We follow \citeauthor{DBLP:journals/corr/abs-2010-10648} (\citeyear{DBLP:journals/corr/abs-2010-10648}) to transcribe the generated images into texts with EasyOCR toolkit and then measure the BLEU \cite{DBLP:conf/acl/PapineniRWZ02} score calculated by SacreBLEU\footnote{https://github.com/mjpost/sacrebleu} \cite{DBLP:conf/wmt/Post18}. To take into account the location of texts within the generated image, we extend the conventional BLEU to Structure-BLEU. This metric first performs OCR on the generated image and the reference image separately. Then, we use the bounding boxes in the generated image and the reference image to calculate intersection over union (IoU) for text matching. Subsequently, we filter out matched text pairs with significantly different positions, specifically those with IoU values below 0.5. Finally, we calculate the BLEU score for the remaining paired texts. For more comprehensive details, please refer to the algorithm provided in the Appendix \ref{appendix:structure-bleu}. Besides, we evaluate the quality of the generated images via structural similarity index measure (SSIM) \cite{DBLP:journals/tip/WangBSS04}, which considers luminance, contrast, and structure to measure the similarity between two images. The comparison between SSIM and BLEU can be found in Appendix \ref{appendix:ssim}.

\begin{figure*}[]
    \centering
    \includegraphics[width=1\linewidth]{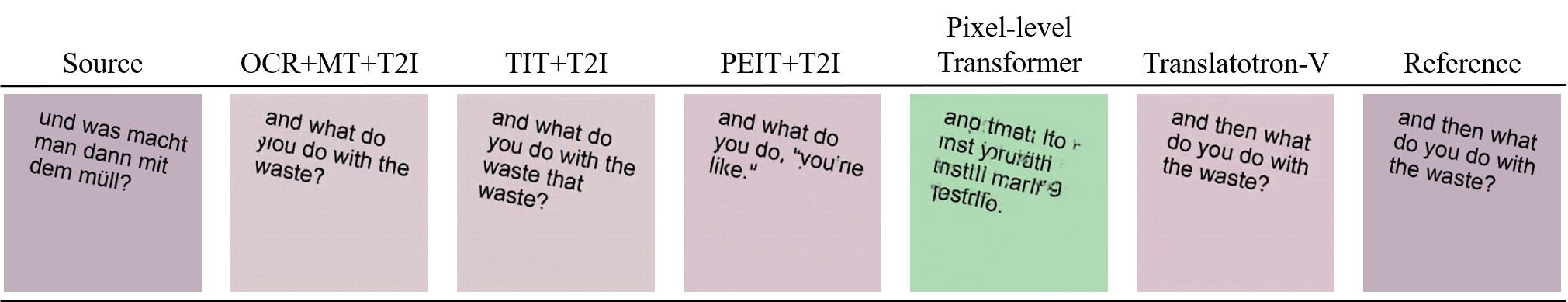}
\caption{Translation examples of different models on the De$\rightarrow$En IIMT task.}

\label{figure:case_study}
\vspace{-0.3cm}
\end{figure*}

\begin{figure}[]
    \centering
    \includegraphics[width=1\linewidth]{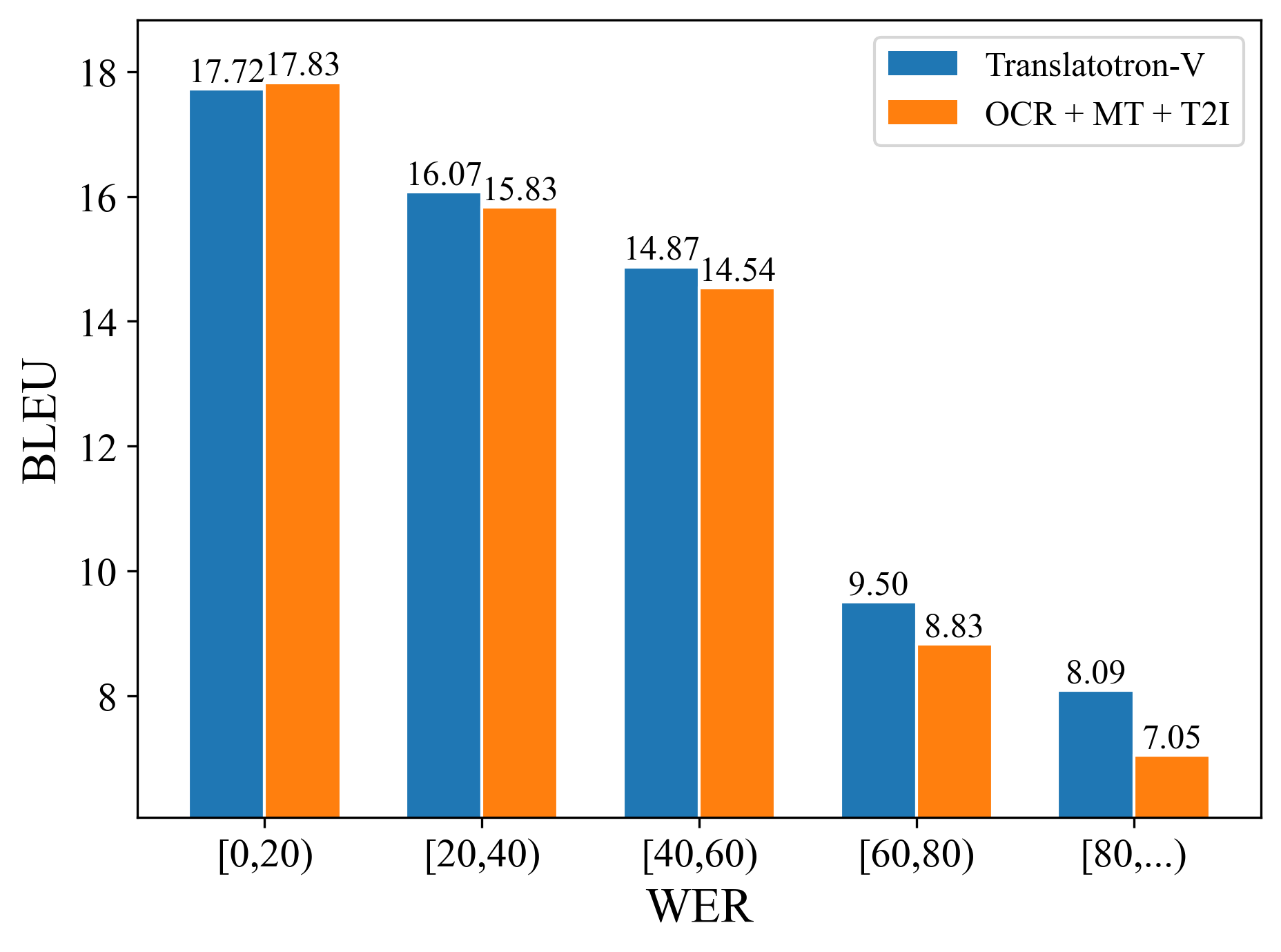}
\caption{BLEU scores on different groups divided according to WER. We compute the WER using the text produced by the OCR toolkit.}
\label{figure:error_prop}
\vspace{-0.2cm}
\end{figure}

\subsection{Results}
Table \ref{tab:de_en_results} shows the overall results of all models on the IWSLT14 De-En datasets, and we have the following observations. First, OCR+MT+T2I exhibits significantly better translation quality compared to TIT+T2I, but its performance of image quality is relatively poor. The underlying reason is that cascading more models may lead to more severe error propagation, thereby affecting the image quality generated by the final T2I model. Second, PEIT+T2I performs worse than other cascaded models, primarily because it heavily relies on the text detection model when handling multi-line text images. Third, the image translation quality generated by Pixel-Transformer is extremely poor, and we observe that it is essentially incapable of producing readable text. This is due to the fact that the search space of image pixel values is extremely large, which poses a challenge to the model's optimization in generating accurate pixel values. Finally, Translatotron-V consistently achieves competitive performance with 70.9\% parameters compared to cascaded baselines and significantly outperforms the pixel-level Transformer.

\subsection{Ablation Study}
To explore the effectiveness of different components, we further compare Translatotron-V with its several variants, as shown in Table \ref{tab:ablation_study}.

1) \textit{w/o gated fusion}. In this variant, we remove the gated fusion mechanism of the image decoder when performing cross-attention. Consequently, the image decoder sequentially performs cross-attention over the image encoder and the target text decoder to update hidden states. The result in Line 2 indicates that this change causes a decline in translation quality, suggesting that the gated fusion mechanism is useful for fusing information from two modalities.

2) \textit{w/o OCR auxiliary task}. When constructing this variant, we remove the OCR auxiliary task during the model training. Upon analyzing Line 3, it becomes evident that this task can empower the model with the ability to perceive text within images, enabling the model to accomplish translation.

3) \textit{w/o knowledge distillation}. We remove the knowledge distillation in this variant. As indicated in Line 4, there is a significant performance drop, which demonstrates that knowledge distillation effectively reduces the difficulty of training.

4) \textit{w/o target text decoder}. In this variant, we remove the target text decoder from our model. The results reported in Line 5 demonstrate a drastic decline in performance. We can confirm that the end-to-end IIMT model imposes a substantial modeling burden. The target text decoder plays a pivotal role in mitigating the burden of achieving alignment between different languages.

\begin{table*}[]
\centering

\resizebox{2\columnwidth}{!}{
\begin{tabular}{lccccccc}
\hline
\multicolumn{1}{l|}{\multirow{2}{*}{\textbf{Model}}}                    & \multicolumn{3}{c|}{\textbf{Fr$\rightarrow$En}}                                                                                                     & \multicolumn{3}{c|}{\textbf{Ro$\rightarrow$En}}                                                                                                    & \multirow{2}{*}{\#Param}       \\
\multicolumn{1}{l|}{}                                          & BLEU $\uparrow$    & Structure-BLEU $\uparrow$ & \multicolumn{1}{c|}{SSIM $\uparrow$}   & BLEU $\uparrow$    & Structure-BLEU $\uparrow$ & \multicolumn{1}{c|}{SSIM $\uparrow$}   &                                \\ \hline
\multicolumn{8}{c}{\textit{Cascaded Models}}                                                                                                                                                                                                                                                                                                                                 \\ \hline
\multicolumn{1}{l|}{OCR+MT+T2I}                            & 21.60                           & 21.58                                  & \multicolumn{1}{c|}{0.7738}                           & 18.34                           & 18.61                                  & \multicolumn{1}{c|}{0.7752}                           & 247M                           \\ 
\multicolumn{1}{l|}{TIT+T2I}                                 & 21.87                           & 21.78                                  & \multicolumn{1}{c|}{0.7801}                           & 18.39                           & 18.30                                  & \multicolumn{1}{c|}{0.7764} & 201M                           \\
\multicolumn{1}{l|}{PEIT+T2I}                                 & 18.51                           & 18.55                                  & \multicolumn{1}{c|}{0.7741}                           & 14.54                           & 14.90                                  & \multicolumn{1}{c|}{0.7704} & 178M                           \\
\hline
\multicolumn{8}{c}{\textit{End-to-end Models}}                                                                                                                                                                                                                                                                                                                                \\ \hline
\multicolumn{1}{l|}{Pixel-level Transformer} & 2.08 & 2.61        & \multicolumn{1}{c|}{0.7753} &1.58  &2.11         & \multicolumn{1}{c|}{0.7696}                           & 162M \\
\hdashline[2pt/2pt]
\multicolumn{1}{l|}{Translatotron-V} & \textbf{22.20} & \textbf{22.17}        & \multicolumn{1}{c|}{\textbf{0.7811}} & \textbf{18.44} & \textbf{18.73}        & \multicolumn{1}{c|}{\textbf{0.7780}}                           & 175M \\ \hline
\end{tabular}
}
\caption{Experimental results on the Fr$\rightarrow$En and Ro$\rightarrow$En IIMT tasks.}
\label{tab:fr_ro_en_results}
\end{table*}

\subsection{Case Study}
Figure \ref{figure:case_study} displays the translation results of different models on the De$\rightarrow$En dataset. We can observe that Translatotron-V generates the correct target image, while the strongest baseline model OCR+MT+T2I missing partial strokes for the word ``\textit{you}'' in the generated images. Besides, Pixel-level Transformer has issues like character omission and artifacts, making it unable to generate correct words. This result reveals that image tokenizer is important for the Translatotron-V.

\subsection{The Effectiveness on Alleviating Error Propagation}
To further investigate the impact of error propagation, we divide the test set of the De$\rightarrow$En dataset into different groups based on the Word Error Rate (WER) of OCR. The higher WER indicates that the image is more difficult to deal with, where potential error propagation is more severe. As illustrated in Figure \ref{figure:error_prop}, the improvements of Translatotron-V over OCR+MT+T2I are more significant with the increase of WER. Thus, we confirm again that our end-to-end model has the potential advantage of alleviating error propagation.

\subsection{Evaluation on Other Language Pairs}
In order to further validate the effectiveness of Translatotron-V, we conduct experiments on two distinct language pairs: French to English (Fr$\rightarrow$En) and Romanian to English (Ro$\rightarrow$En). We also use the previously-described data synthesis method to convert the IWSLT17 Fr-En and Ro-En datasets \cite{DBLP:conf/iwslt/CettoloFBNSSYF17} into IIMT datasets. As shown in Table \ref{tab:fr_ro_en_results}, Translatotron-V still achieves competitive performance compared to cascaded models and significantly outperforms the Pixel-level Transformer across different language pairs.

\section{Conclusion}
In this work, we have proposed Translatotron-V, which is the first end-to-end IIMT model capable of generating RGB images and achieving comparable performance to the cascaded model with only 70.9\% of parameters. In addition to an image encoder and an image decoder, Translatotron-V is equipped with a target text decoder and an image tokenizer, which are used to alleviate the modeling burden and prevent the model from directly predicting pixels, respectively. Moreover, we present a two-stage training framework to assist the model in learning alignment across modalities and languages. Furthermore, we introduce an evaluation metric, Structure-BLEU, which considers text location information to evaluate the quality of translations within the image. Experimental results demonstrate the effectiveness of our proposed model and training framework.

In the future, we are interested in training models using only parallel images, which is important when texts within the image are not available.

\section*{Limitations}
Currently, the quality of generated target images depends on the quality of the image tokenizer. However, in our experiments, we find that it sometimes generates incorrect words, which may be due to its training using only images without explicitly considering linguistic information. Meanwhile, Translatotron-V does not exhibit a speed advantage over the cascaded model. This is due to the reason that visual token sequences are still much longer than text sequences, and both cascaded and end-to-end models need to spend most of their time decoding the image. A promising direction is to find coarser visual tokens with a shorter sequence length without degrading the quality of the generated images. Furthermore, the synthetic dataset is still not realistic enough. However, acquiring IIMT data from the real world is very challenging, how to create a more realistic IIMT dataset is also an important direction.

\section*{Acknowledgments}
The project was supported by 
National Natural Science Foundation of China (No. 62036004, No. 62276219), and the Public Technology Service Platform Project of Xiamen (No. 3502Z20231043). We also thank the reviewers for their insightful comments.
\bibliography{anthology,custom}
\bibliographystyle{acl_natbib}

\appendix
\section{The effectiveness of the training objective with balancing coefficients}
\label{appendix:weight_balance}

\begin{figure}[h]
    \centering
    \includegraphics[width=1\linewidth]{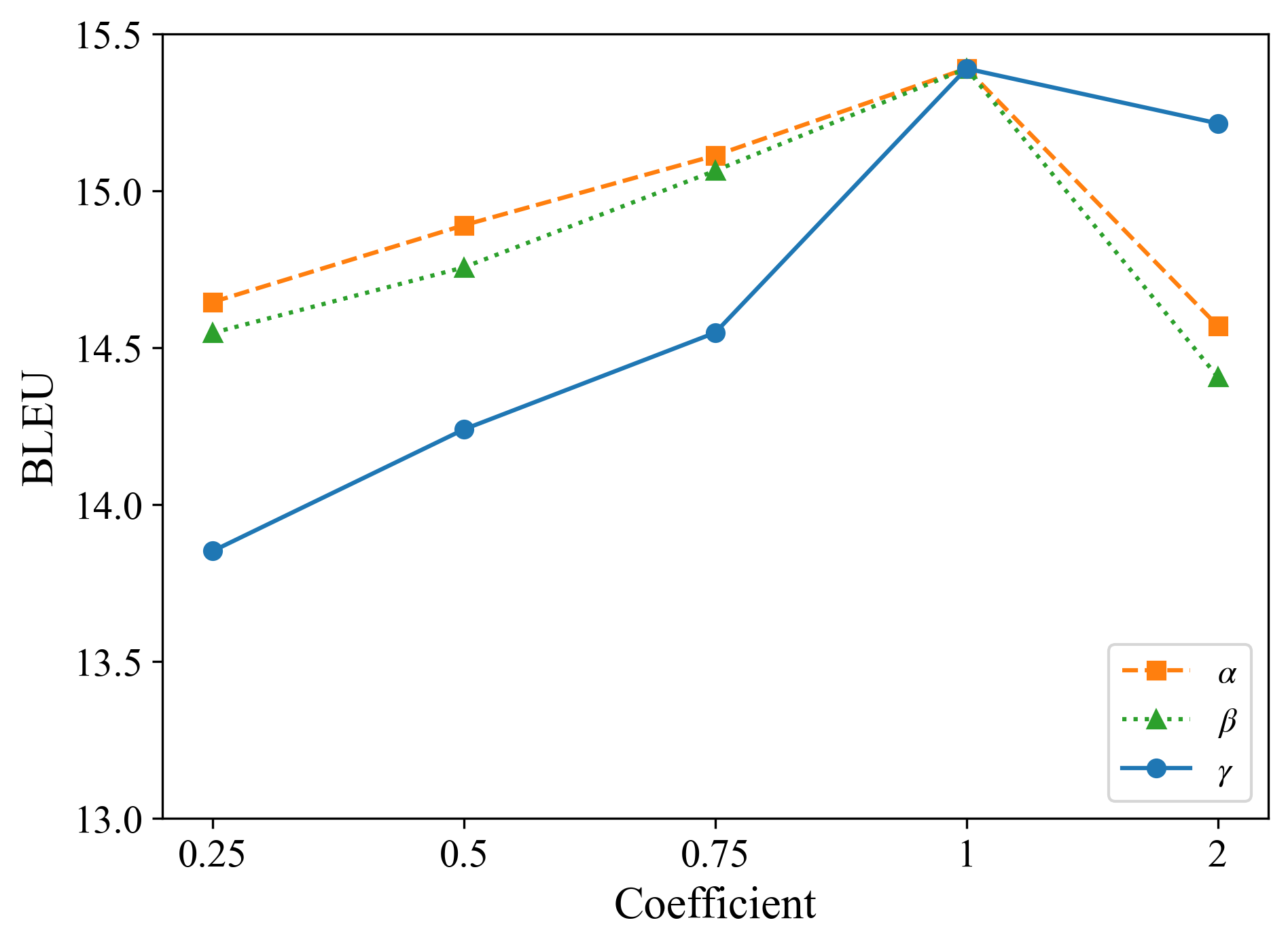}
\caption{BLEU scores with different loss function coefficients.}
\label{figure:loss_weight}
\end{figure}

To explore the impact of different weights of auxiliary task losses on model performance, we modify the training objective at the second stage as follows:
\begin{equation}
    \mathcal{L}_2 = \mathcal{L}_{iimt} + \alpha \mathcal{L}_{ocr} + \beta \mathcal{L}_{tit} + \gamma \mathcal{L}_{kd}.
\end{equation}
where $\alpha$, $\beta$, and $\gamma$ are the coefficient to control OCR auxiliary task loss $\mathcal{L}_{ocr}$, TIT auxiliary task loss $\mathcal{L}_{tit}$, and knowledge distillation loss $\mathcal{L}_{kd}$, respectively.

Due to the high cost of grid search, when adjusting a specific coefficient, all other coefficients will be set to 1. As shown in Figure \ref{figure:loss_weight}, when all coefficients are set to 1, the model performs optimally. This result suggests that these tasks may be equally important and highly correlated. Adjusting a particular coefficient could lead the model to focus more on or neglect one task, which might not be beneficial if all tasks are equally important. This also implies that our approach does not require carefully adjusting the coefficients for different training objectives.

\begin{figure*}[h]
    \centering
    \includegraphics[width=0.9\linewidth]{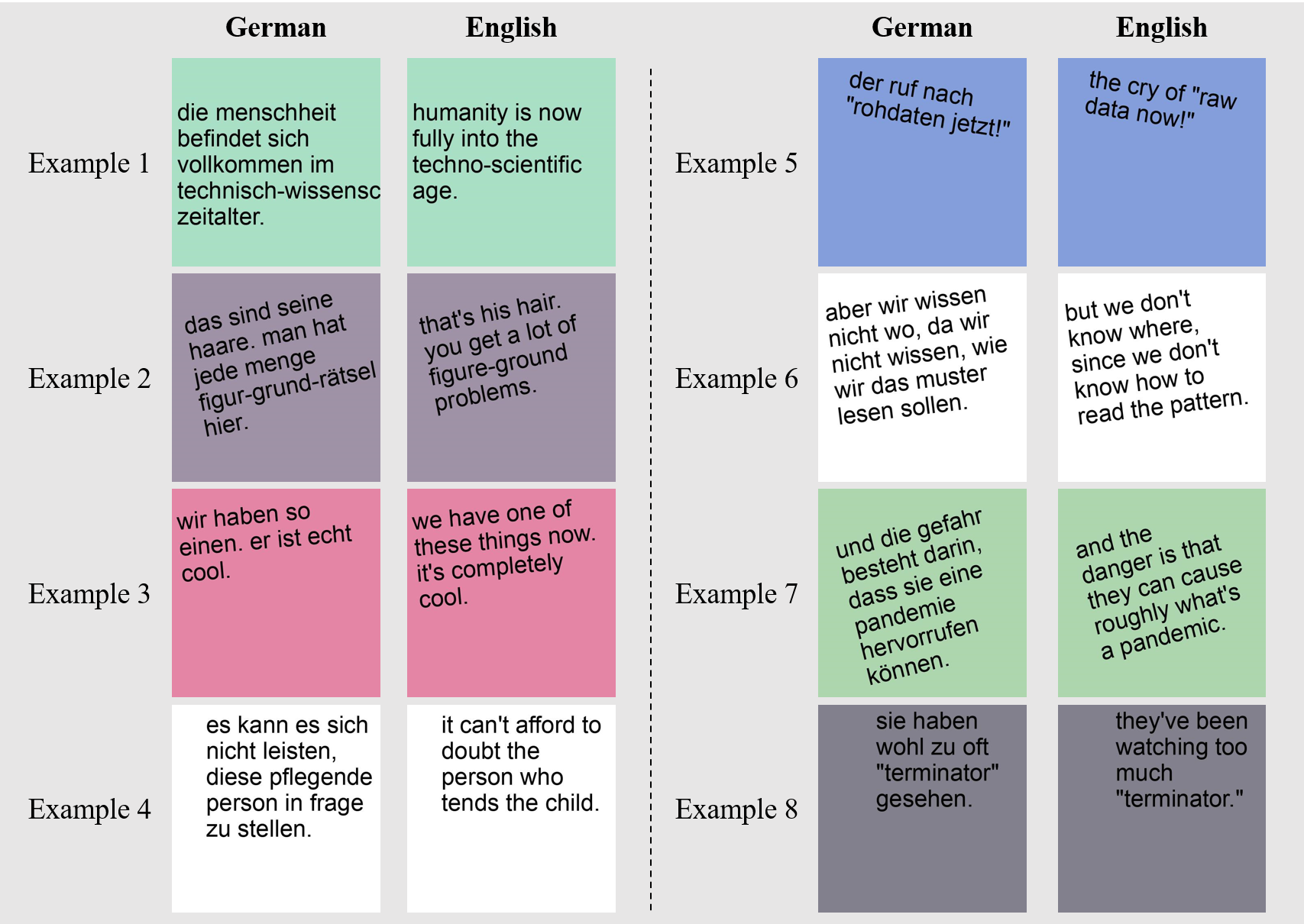}
\caption{Several examples of our synthetic data using IWSLT14 De-En.}
\label{figure:data_example}
\end{figure*}

\begin{figure*}[h]
    \centering
    \includegraphics[width=1.0\linewidth]{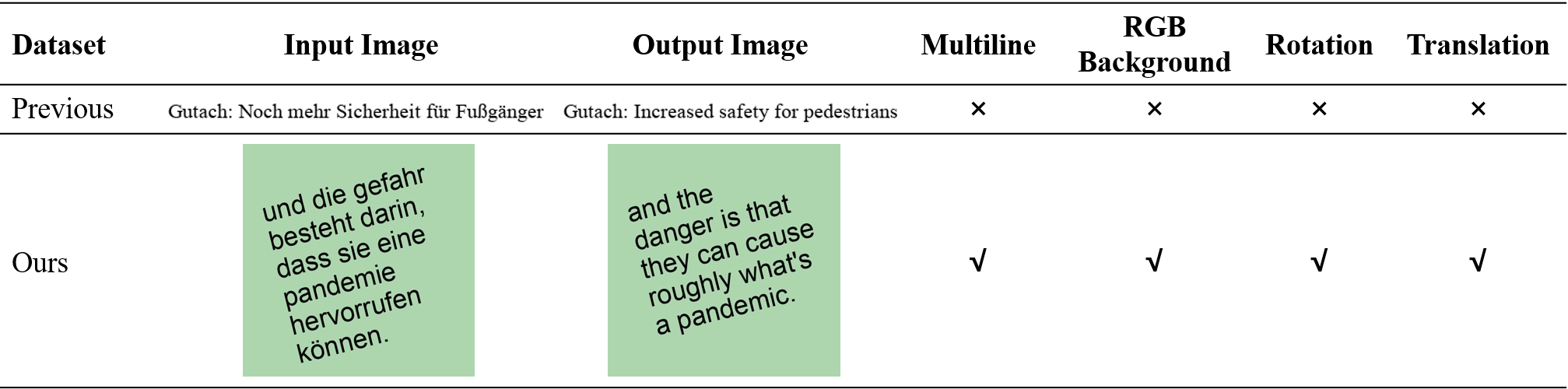}
\caption{Comparison of our IIMT data with other IIMT data used in previous work \cite{DBLP:journals/corr/abs-2010-10648,DBLP:conf/emnlp/TianLLGW23}.}
\label{figure:data_compare}
\end{figure*}

\section{Data Examples}
\label{appendix:data_examples}
In Figure \ref{figure:data_example}, we present several data examples for our synthetic data. We also show the difference between our IIMT data and previous IIMT data in Figure \ref{figure:data_compare}. It can be observed that our data is more complex than the data used in previous work.

\section{Baseline Architecture Details}
\label{appendix:baseline_details}
In this section, we provide a detailed description of the architectures of two baselines: OCR+MT+T2I and TIT+T2I.

\textit{OCR+MT+T2I}. This baseline cascades three models: an OCR model, an MT model, and a T2I model. 
First, we follow \citeauthor{DBLP:conf/aaai/LiLC0LFZ0W23} (\citeyear{DBLP:conf/aaai/LiLC0LFZ0W23}) to construct the OCR model, which consists of a ViT-B encoder and a Transformer decoder. The decoder uses the settings of Transformer Base, which contains 6 layers, 8 attention heads, 512-dimensional hidden states, and 2048 feed-forward hidden states. Second, we use the standard Transformer base\cite{DBLP:conf/cvpr/HeZRS16} as the architecture of the MT model. Third, the T2I Model includes a Transformer-based text encoder, a ResNet-based image encoder \cite{DBLP:conf/cvpr/HeZRS16}, an image decoder, and an image tokenizer. Both the text encoder and image decoder utilize the same settings as Transformer base. Additionally, the image encoder adopts the ResNet50 architecture, and the image tokenizer follows the configuration of our model.
It's worth noting that our teacher model has the same structure as this T2I Model, except for reducing the hidden states from 512-dimensional to 384-dimensional.

\textit{TIT+T2I}. We construct this baseline by cascading the TIT model and the T2I model, where the TIT model consists of an image encoder, a source text decoder, and a target text decoder. The image encoder uses the same architecture as ViT-B, and the configuration of the source text decoder and target text decoder is consistent with Transformer Base. 

\textit{PEIT+T2I}. This baseline replaces the multi-line TIT model in TIT+T2I with PEIT \cite{DBLP:conf/acl/ZhuLLX23}. To ensure a fair comparison, we reimplement the model, scale its size to match our model, and do not use additional data during its multi-stage training.

\begin{figure*}[h]
    \centering
    \includegraphics[width=0.9\linewidth]{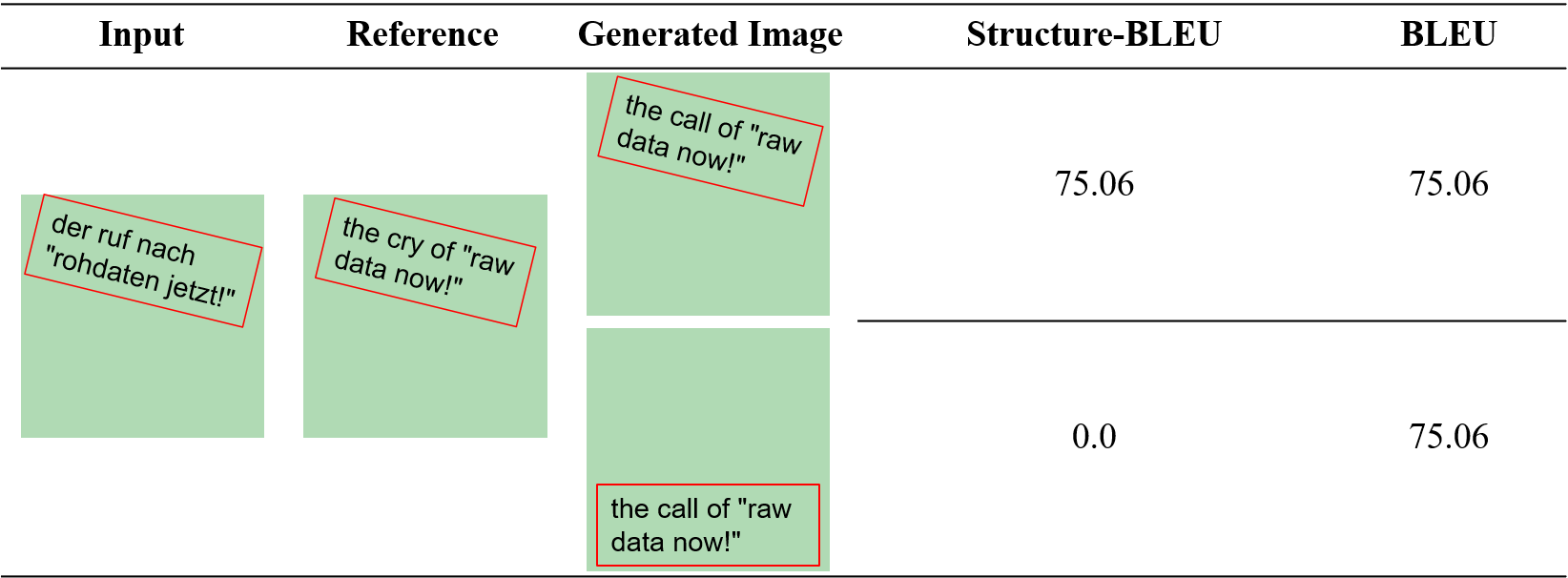}
\caption{Comparison of Structure-BLEU and BLEU. The bounding box of the text is drawn with red lines.}
\label{figure:s-bleu_vs_bleu}
\end{figure*}

\section{Structure-BLEU}

In Algorithm \ref{algorithm:structure-bleu}, we provide the detailed calculation process of Structure-BLEU. Furthermore, as shown in Figure \ref{figure:s-bleu_vs_bleu}, we also provide an example to demonstrate the difference between Structure-BLEU and BLEU. We can observe that Structure-BLEU takes into account the positional information of the text in the generated image, which is crucial for user experience in real-world scenarios.

\label{appendix:structure-bleu}
\begin{algorithm}
\caption{Structure-BLEU}
\label{algorithm:structure-bleu}
\begin{algorithmic}[1]
\REQUIRE Generated image $\hat{\mathbf{t}}$, reference image $\mathbf{t}$

\STATE $\mathcal{R} \leftarrow \text{OCR}(\mathbf{t})$ \COMMENT{Set of reference texts and bounding boxes}
\STATE $\mathcal{H} \leftarrow \text{OCR}(\hat{\mathbf{t}})$ \COMMENT{Set of generated texts and bounding boxes}
\STATE $\mathcal{M} \leftarrow \{\}$ \COMMENT{Set of matched text pairs}

\FOR{each $h \in \mathcal{H}$}
    \STATE $m \leftarrow \text{None}$
    \STATE $s \leftarrow 0$
    \FOR{each $r \in \mathcal{R}$}
        \STATE $\hat{s} \leftarrow \text{IOU}(h, r)$
        \IF{$\hat{s} > s$}
            \STATE $s \leftarrow \hat{s}$
            \STATE $m \leftarrow (h, r)$
        \ENDIF
    \ENDFOR
    
    \IF{$s \geq 0.5$}
        \STATE $\mathcal{M} \leftarrow \mathcal{M} \cup \{m\}$
    \ENDIF
\ENDFOR
\RETURN BLEU($\mathcal{M}$)
\end{algorithmic}
\end{algorithm}

\begin{figure*}[h]
    \centering
    \includegraphics[width=0.8\linewidth]{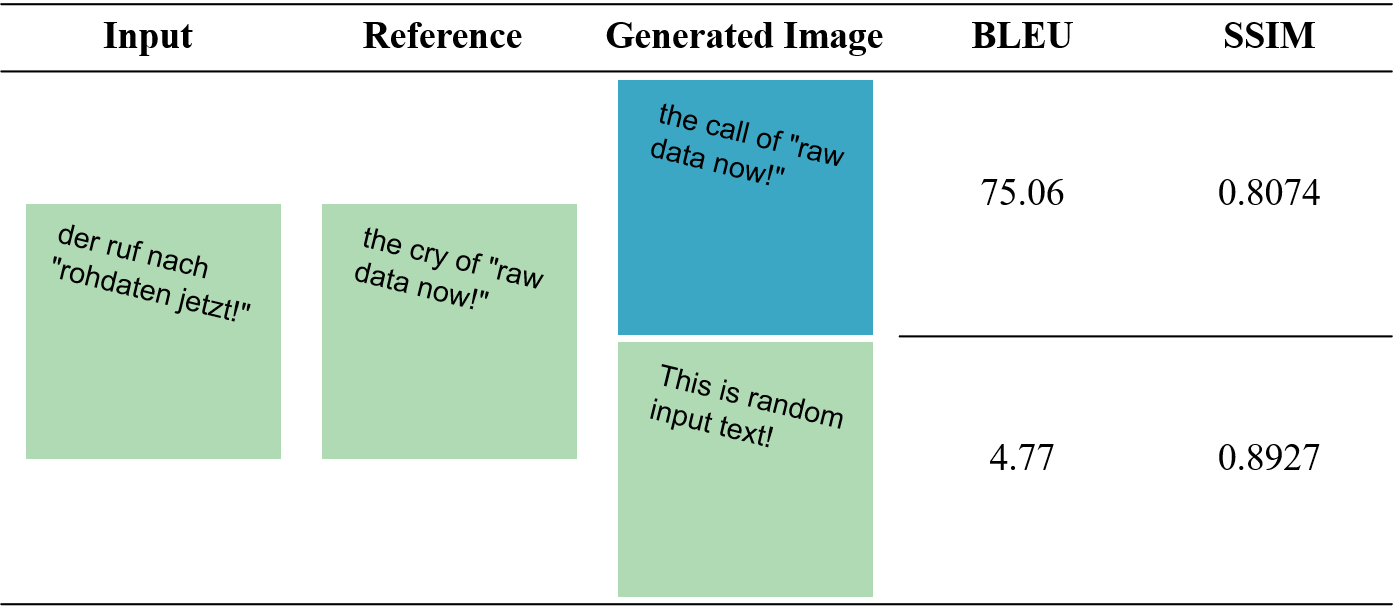}
\caption{Comparison of BLEU and SSIM.}
\label{figure:bleu_vs_ssim}
\end{figure*}

\section{Comparison of BLEU and SSIM}
In this section, we provide examples illustrating the differences in focus between SSIM and BLEU evaluations. As shown in Figure \ref{figure:bleu_vs_ssim}, even when the text in the generated image is substantially incorrect, SSIM still yields a relatively high score. The underlying reason is that SSIM is used to evaluate the visual similarity between the generated image and the reference image, considering aspects such as brightness, contrast, etc. It focuses not only on the image of text regions but also on the image of non-text regions. However, BLEU is employed to evaluate the fine-grained information in the image of text regions, such as the glyph of the character. The two metrics have different focuses and complement each other. Therefore, the model may perform well in generating visual information, while it may struggle with generating fine-grained text details, which will lead to a significant difference in BLEU scores but relatively close in SSIM scores.

\label{appendix:ssim}

\end{document}